\documentclass{article}

\usepackage{PRIMEarxiv}

\usepackage[utf8]{inputenc} 
\usepackage[T1]{fontenc}    
\usepackage{hyperref}       
\usepackage{url}            
\usepackage{booktabs}       
\usepackage{amsfonts}       
\usepackage{nicefrac}       
\usepackage{microtype}      
\usepackage{lipsum}
\usepackage{fancyhdr}       
\usepackage{graphicx}       
\graphicspath{{media/}}     

\usepackage{times}
\usepackage{latexsym}

\usepackage[T1]{fontenc}
\usepackage[utf8]{inputenc}
\usepackage{microtype}
\usepackage{inconsolata}
\usepackage{multirow}
\usepackage{graphicx}
\usepackage{amsfonts}
\usepackage{amsmath}
\usepackage{amsthm}
\usepackage{amssymb}
\usepackage{amsmath}
\usepackage{natbib}

\title{Grounding Spatial Relations in Text-Only Language Models
}

\author{
  Gorka Azkune, Ander Salaberria, Eneko Agirre \\
  HiTZ Basque Center for Language Technologies - IXA NLP Group \\
  University of Basque Country (UPV/EHU) \\
  Donostia, Basque Country, Spain\\
  \texttt{\{gorka.azcune, ander.salaberria, e.agirre\}@ehu.eus}
}

\begin{document}
\maketitle

\begin{abstract}
This paper shows that text-only Language Models (LM) can learn to ground spatial relations like \emph{left of} or \emph{below} 
if they are provided with explicit location information of objects and they are properly trained to leverage those locations. We perform  experiments on a verbalized version of the Visual Spatial Reasoning (VSR) dataset, where images are coupled with textual statements which contain real or fake spatial relations between two objects of the image. We verbalize the images using an off-the-shelf object detector, adding location tokens to every object label to represent their bounding boxes in textual form.  Given the small size of VSR, 
we do not observe any improvement when using locations, but pretraining the LM over a synthetic dataset  automatically derived by us improves results significantly when using location tokens. We thus show that locations allow LMs to ground spatial relations, with our text-only LMs outperforming Vision-and-Language Models and setting the new state-of-the-art for the VSR dataset. Our analysis show that our text-only LMs can generalize beyond the relations seen in the synthetic dataset to some extent, learning also more useful information than that encoded in the spatial rules we used to create the synthetic dataset itself.
\end{abstract}


\section{Introduction}
Spatial relations like \emph{left of} or \emph{on top of} 
can be naturally grounded to images. Thus, Vision-and-Language Models (VLM) seem the most suitable option to ground the textual form to real world concept usage. However, general-purpose VLMs such as CLIP \citep{pmlr-v139-radford21a}, VisualBERT \citep{li2019visualbert}, LXMERT \citep{tan2019lxmert} or ViLT \citep{kim2021vilt} have been shown to struggle to ground spatial relations properly \citep{liu2022things, liu2022visual}. The situation is even worse for text-only LMs, which lag behind VLMs for spatial grounding \citep{liu2022things}.

\begin{figure}[t]
\centering
\includegraphics[width=11cm]{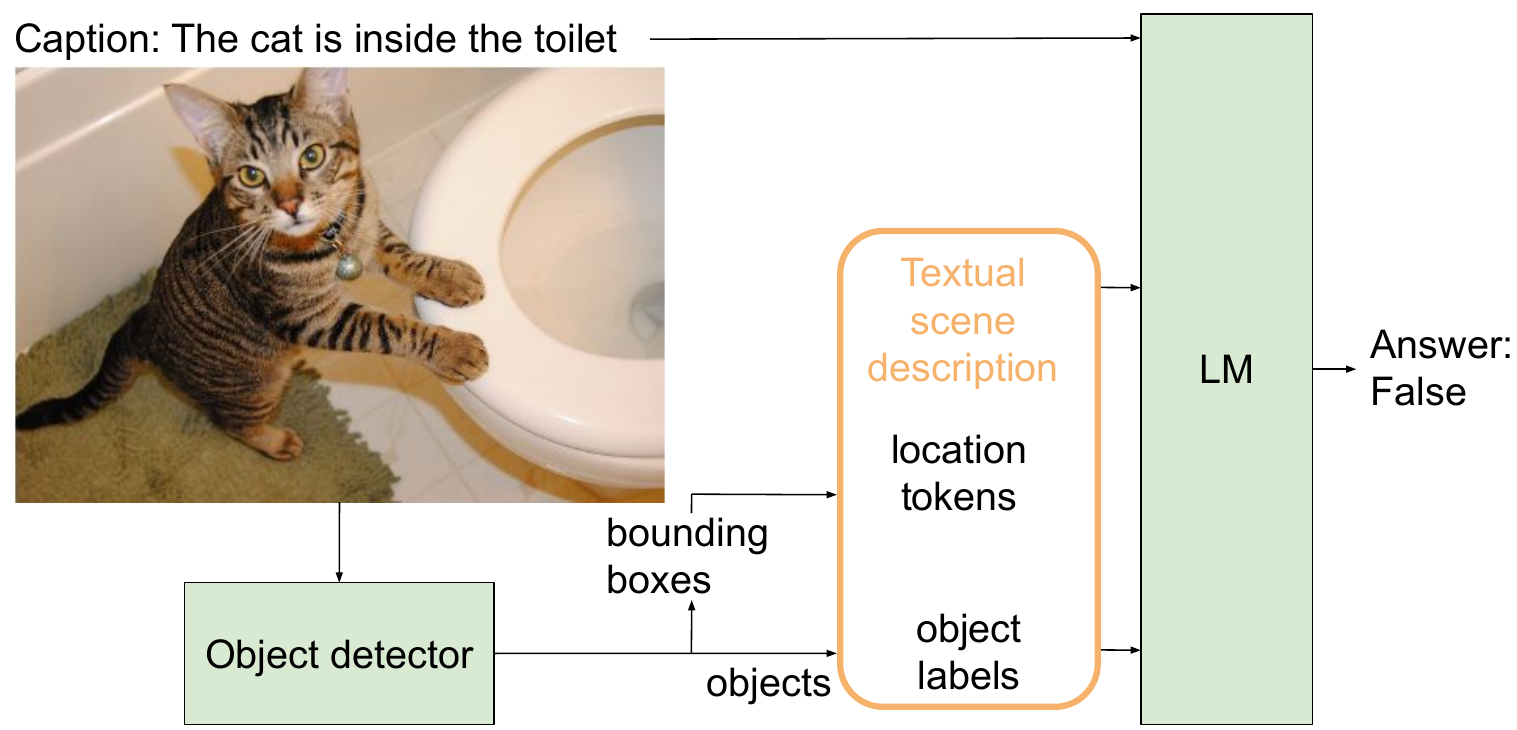}
\caption{Given an image and a caption with a spatial relation, the task in VSR is to output whether the caption is true for the image. We propose a text-only alternative of the dataset, where an off-the-shelf object detector returns the labels and locations (derived from the bounding boxes), which are used as the textual description of the scene depicted in the image. The description and caption are input to a LM, to test its spatial grounding capabilities.} \label{fig:system}
\end{figure}

Spatial grounding and reasoning are very interesting for text-only tasks, as shown by various works \citep{liu2022things, mirzaee2021spartqa, mirzaee-kordjamshidi-2022-transfer}. One alternative to solve those text-only tasks would be using VLMs and feed them only with textual inputs. However, some researchers already identified that the language used to train those VLMs is not as rich and varied as the language used for text-only tasks \citep{tan2020vokenization}, which hinders the potential of VLMs for text-only tasks.

In this paper, we explore another avenue and we focus on spatial grounding for text-only LMs. Following the current trend of translating visual information into textual information \citep{yang2022empirical, zeng2022socratic, wang2022language, liu2022deplot}, we propose to use textual tokens in a novel way to represent real-world scenes and leverage pretrained LMs.
More concretely, we propose to use location tokens to represent the positions and spatial extent of objects in a scene. Our hypothesis is that those location tokens offer a way to ground spatial relations in the LM. 

To validate that hypothesis, we run experiments on a verbalized version of the multimodal Visual Spatial Reasoning (VSR) dataset \citep{liu2022visual}. The dataset contains image-caption pairs, where the caption mentions a spatial relation between two objects of the image, plus a true/false label, depending if the caption is true for the image. To approach this task with a text-only LM, we use an off-the-shelf object detector, which returns object labels and their bounding boxes (BB). We convert the BB coordinates to four location tokens. We prepend the location tokens to the corresponding object label (e.g. \emph{cat}), and build a \textbf{textual scene description} that represents the contents and locations in a given scene (Figure \ref{fig:system}). Then, we only concatenate the provided caption with the aforementioned textual scene description and train a LM for binary classification (Figure \ref{fig:system}). This way, we can test the spatial grounding capabilities of a text-only LM. 

As a result of our experiments we show that:
\begin{enumerate}
\item 
Location tokens are effective to ground spatial relations, as shown by the positive results of our model. 
\item 
The training set of VSR is too small for learning how to ground spatial relations to locations, but an automatically produced synthetic dataset of spatial relations allows to do so, while a LM without locations fails. 
\item 
The LMs trained on the synthetic dataset can generalize to some extent to spatial relations that have not been observed in the synthetic data. Specially interesting is to see the performance boost for relations that require depth information.
\item
Our text-only LMs outperform baseline VLMs for VSR, obtaining the best results for the VSR task to date. 
\item Our text-only LMs clearly outperform a rule-based baseline, showing that the LMs learn more information than that encoded in the manually defined spatial rules.
\end{enumerate}

Our code, models and datasets are freely available\footnote{https://github.com/gazkune/SpatialLM}.

\section{Related work}
Some authors suggest that grounding is one of the key elements to bring human-like language understanding \citep{bender2020climbing}. However, grounding covers a great diversity of techniques, modalities and concepts \citep{van2015communication, laflaquiere2018discovering}. Thus, this paper is focused on spatial relations and their grounding. In that sense, there are two major domains related to this paper: how spatial grounding can be evaluated (Section \ref{sec:spatial-grounding}), and how spatial information is represented in current deep learning models, covering VLMs - which are the current paradigms of how to ground text on visual data - and text-only LMs (Section \ref{sec:spatial-encoding}).

\subsection{Datasets for spatial grounding}
\label{sec:spatial-grounding}
The spatial commonsense knowledge of current LMs and VLMs is evaluated from different angles. For example, \citep{bagherinezhad2016elephants, elazar2019large} focus on the acquired commonsense knowledge of models about object scales, e.g. do they know that a person is bigger than an ant? In that sense, they do not provide a specific scene context, but rather ask about generic object scale relations, so the dataset they provide is not useful for our work.  

Some other authors, \citep{collell2018acquiring, elu2021inferring} propose datasets and methods to generate bounding boxes from textual descriptions. Although the evaluation approach is suitable to test spatial grounding, they focus on implicit spatial relations, whereas our focus is on explicit relations. Thus, the proposed datasets are not suitable for our analysis. 

With the objective of evaluating both object scales and spatial relations, a recent work provides new unified datasets \citep{liu2022things}. As the objective of such work is to evaluate whether VLMs learn more spatial commonsense than LMs, the datasets are purely textual, so they do not provide any means to ground spatial relations (they assume the grounding occurs in a previous training process) and hence, they are not useful for our work. Interestingly, authors find that VLMs, and more concretely text-to-image systems, perform much better than text-only LMs. 

There are other ways to test the spatial inference and reasoning capabilities of models, though. CLEVR was one of the pioneering works on testing compositional language and elementary visual reasoning \citep{johnson2017clevr}. Using 3D rendered images of simple objects such as spheres, cones and cubes, different questions are generated automatically. A model has to process the image and the question to provide an answer. Although CLEVR can be used to test spatial grounding, it has two major drawbacks for the work presented in this paper: i) questions not only cover spatial grounding but some other concepts such as compositional language and attribute identification, and ii) spatial relations are limited to four, i.e. \emph{left}, \emph{right}, \emph{behind} and \emph{in front}. The natural extension of CLEVR is GQA \citep{hudson2019gqa}, which shares similar ideas but it is built on natural images. Although spatial grounding is very important for this task, compositional language is also evaluated. As both dimensions appear together, we believe this dataset is not the best option for our purposes.

In the text-only scenario, SpartQA provides another synthetic question-answering dataset (there is also a subset annotated by humans). Given a textual story (a spatial description of a scene using explicit relations), a model has to answer some spatial questions about that scene. The task is specially focused on spatial reasoning capabilities, such as transitivity, and it does not provide any means to ground spatial relations, as its target is the reasoning process. Recently, similar datasets haven been proposed as an extension and improvement of SpartQA \citep{mirzaee-kordjamshidi-2022-transfer}.

In this paper, we use the recent Visual Spatial Reasoning (VSR) dataset \citep{liu2022visual} to evaluate the spatial grounding capabilities of text-only LMs. VSR has been designed to test spatial grounding capabilities, covering 65 different spatial relations over natural images collected from COCO \citep{lin2014microsoft}. Given an image, they provide a caption which describes a spatial relation between two of the objects that appear in the image. That relation can be real or fake, and that is precisely what the model has to infer, i.e. whether the caption is correct respect to the given image. The dataset is fully annotated by humans. Given its features, we believe VSR is a good candidate to evaluate spatial grounding in LMs and thus, we use it in our experiments. However, as text-only LMs cannot process images, we propose a way to verbalize those images and run meaningful experiments.

\subsection{Encoding spatial information}
\label{sec:spatial-encoding}
The most successful VLMs today are based on multimodal transformers \citep{tan2019lxmert, kim2021vilt}. Although architectures may vary, the basic idea is to input the models with textual tokens and visual features. As transformers are feed-forward networks, they do not consider the order of the inputs, and thus, positional encodings are used to represent, for example, word order \citep{NIPS2017_3f5ee243}. A similar idea is used also for visual features. LXMERT \citep{tan2019lxmert}, for instance, uses the $x_0, y_0, x_1, x_2, W, H$ coordinates of a bounding box for a given visual feature, projects them linearly and sums it to the visual feature itself before inputting it to the transformer. Alternatively, ViLT \citep{kim2021vilt}  does not use any object detector, but works directly on image patches. They use positional embeddings to represent the order of those patches in the image, very similar to the positional embeddings of textual tokens. 

Regarding text-only LMs, to the best of our knowledge, \citep{patel2022mapping} is the only work where scenes are represented with textual tokens on which spatial grounding and reasoning can be performed. More concretely, they propose to create grid-like structures with textual tokens inside the vocabulary of the LM. Their proposal is interesting, but it is limited to toy experiments, since they can only represent \textit{small} scenes and six spatial relations: \textit{left}, \textit{right}, \textit{up}, \textit{down}, \textit{top} and \textit{bottom}. In contrast, with our approach we cover complex scenes depicted in natural images and 23 spatial relations (Table \ref{tab:std-relations}).

\section{The VSR dataset}

\begin{figure}[t]
\centering
\includegraphics[width=11cm]{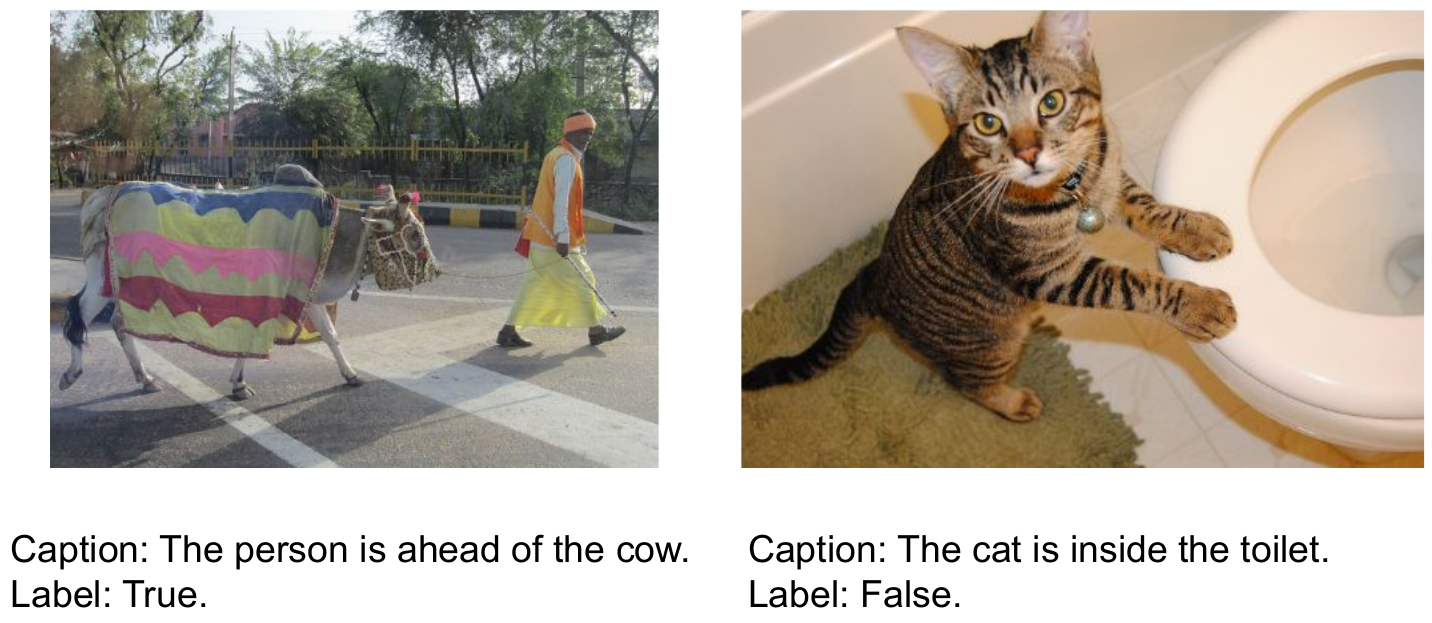}
\caption{Two examples extracted from the VSR dataset.} \label{fig:vsr-examples}
\end{figure}

The VSR dataset contains natural image-text pairs to test the spatial grounding capabilities of machine learning models. As can be seen in Figure \ref{fig:vsr-examples}, a textual description of an image is provided, where the spatial relation of two objects is explicitly described. The spatial relation can be true or false. To solve the task properly, models have to be able to ground around 65 different spatial relations, which are grouped in 7 categories: adjacency, directional, orientation, projective, proximity, topological and unallocated. 

The dataset has two splits: the \emph{random} split and the \emph{zero-shot} split. The later is designed such that train/dev/test sets have no overlapping concepts and force models to learn concepts and the relations in a compositional way instead of memorizing co-occurrence statistics of the two. However, it is smaller than the random split, which has a total of 10,119 examples. The zero-shot split has 6,430 image-text pairs in total. 

According to the experiments performed in the VSR dataset by authors \citep{liu2022visual}, the best VLMs are far from human performance. While humans obtain an accuracy of 95.4 for both splits, the best model for the random split, i.e LXMERT \citep{tan2019lxmert}, is around 70.1 and it performs worse in the zero-shot split (63.0). This performance gap between humans and VLMs shows that there is still much work to do to better ground spatial relations.

\section{Learning to ground spatial relations in text-only LMs}

In this paper, we propose to ground spatial relations in LMs introducing the concept of \emph{location tokens}. These location tokens use numbers that are already in the vocabulary of the LM. Thus, we can represent any scene, using four location tokens to represent the position and the spatial extension of an object and combining them with the object name (and any other object attribute). This textual scene representation allows LMs to relate spatial relations like \textit{left of} with specific arrangement of location tokens, providing a way to ground those relations.

To test our hypothesis, we verbalize the VSR dataset and use it for training and evaluation. As Figure \ref{fig:system} shows, we approach the problem stated in VSR in the following way: (i) we obtain textual scene descriptions using an object detector, (ii) we include in that description the location tokens derived from the object bounding boxes, (iii) we concatenate the caption and the textual scene description and input it to the LM, (iv) we fine-tune the LM on that input for binary classification. We also offer the possibility to previously train the LM in our Synthetic Spatial Training Dataset.

\subsection{Textual scene descriptions}
Given that VSR is a visio-linguistic dataset, the scene is defined by an image. We convert that scene to a textual description using a state-of-the-art object detector, VinVL \citep{zhang2021vinvl}, which given an image, produces a list of objects with their name, attributes and bounding boxes. More concretely, an object detected by VinVL is represented as $O = \{ name, attr_1, \ldots, attr_n, BB \}$, where $BB \in \mathbb{R}^4$ are the $x_0, y_0, W, H$ coordinates of the bounding box. 

\begin{figure}[t]
\centering
\includegraphics[width=11cm]{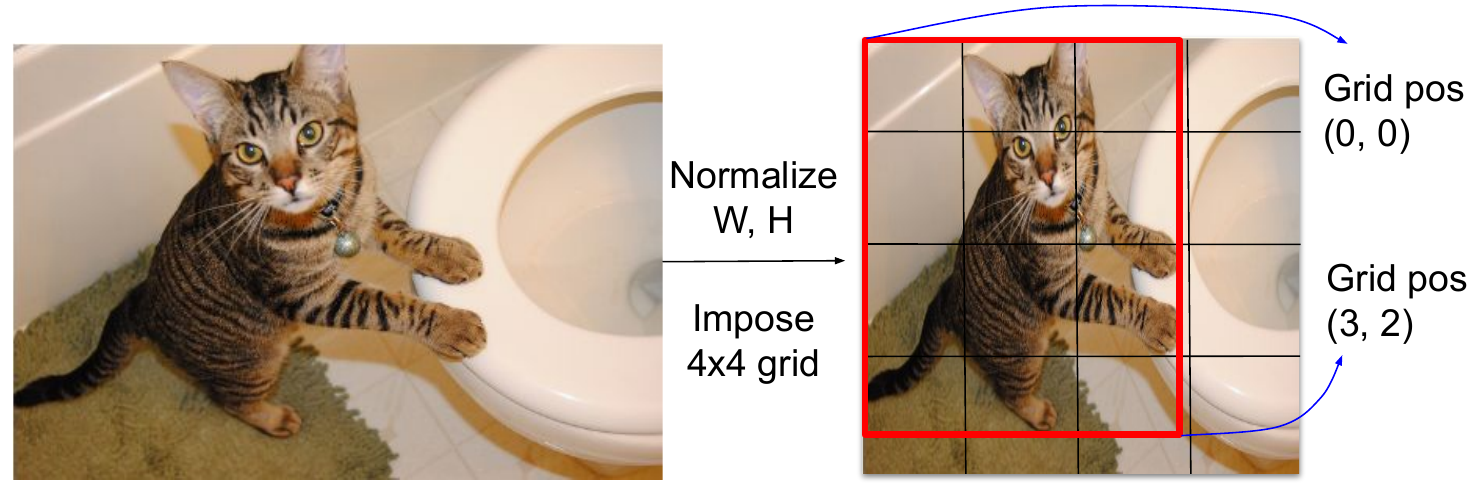}
\caption{An illustrative example of how BB coordinates are converted to location tokens. In this case, with a grid size of $4 \times 4$, the location tokens for cat (red box) are \emph{0 0 3 2}.} \label{fig:location-tokens}
\end{figure}

To convert those BBs to location tokens, we follow this procedure (Figure \ref{fig:location-tokens}): (i) normalize the image's width and height in the $[0, 1]$ range, (ii) divide the image in a regular grid of size $(G \times G)$, and (iii) find the grid cells for the BB coordinates $(x_0, y_0, x_1, y_1)$ which we call $(\hat{x}_0, \hat{y}_0, \hat{x}_1, \hat{y}_1)$, i.e. discrete coordinates. Those discrete coordinates (after tokenization of the corresponding strings) are the location tokens. As a result, for every object detected, we get a sequence of four location tokens or discrete coordinates. Thus, our textual scene description $Descr(S)$ is a sequence of textual objects $\{O_0, O_1, \ldots O_N\}$, where each object is a string of the form: $O_i = \{ \hat{x}_0^i, \hat{y}_0^i, \hat{x}_1^i, \hat{y}_1^i, name_i \}$. Notice that VinVL also returns a list of attributes for every object. Unless stated otherwise, we discard those attributes in the textual scene description.

For the VSR task, we produce textual descriptions for all the images, concatenate them with the captions provided in the dataset and input it to the LM. Using positional embeddings, the LM can learn to interpret properly the order of location tokens and their correspondence with the object names. For example, for the image in Figure \ref{fig:location-tokens}, the textual description of the object \emph{cat} is: \emph{0 0 3 2 cat}. Assuming that our grid size $G=4$, this is interpreted as having a cat covering the left part of the image. We would do similarly for all the objects of the image to build our textual scene description.

Notice that for VSR, the textual scene descriptions are derived from images. But in the general case, we could derive them from other modalities like graphs or text. For instance, given a natural textual description of a scene (e.g. "a cat is on top of a table"), textual scene descriptions with location tokens could be obtained. However, as we could not find any suitable dataset for those cases, we leave them out of the scope of this paper (see Section \ref{sec:conclusions}).

\subsection{The Synthetic Spatial Training Dataset}
\begin{table*}
\centering
\begin{tabular}{cc}
\toprule
\textbf{Category} & \textbf{Spatial Relations}\\
\midrule
\multirow{2}{*}{Object position in the image} & top left, bottom left, left, top right, \\ & bottom right, right, top, bottom, center \\
\midrule
Object size comparison & wider, narrower, taller, shorter, larger, smaller \\
\midrule
\multirow{2}{*}{Two object positional relations} & surrounding, inside, left of, above, right of, \\ & below, overlapping, separated \\
\bottomrule
\end{tabular}
\caption{The 23 relations in our Synthetic Spatial Training Dataset organized in three categories.  }
\label{tab:std-relations}
\end{table*}

\begin{figure}
    \centering
    \includegraphics[width=8cm]{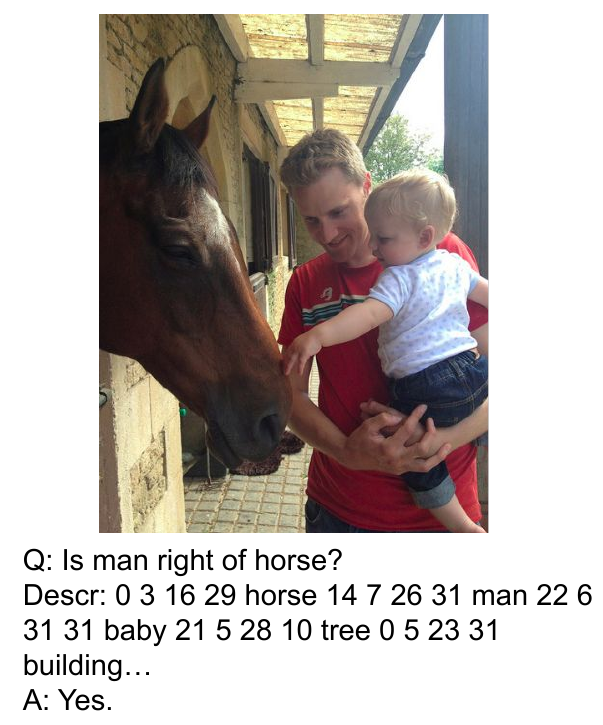}
    \caption{An example of the SSTD validation set generated from the image, which includes question (Q), description (Descr) and answer (A), but not the image itself. Description partially shown, as it comprises 44 objects. Location tokens are discrete grid coordinates 
    of the BB, e.g. $(0, 3)$ and $(16, 29)$ for horse. } 
    \label{fig:std-example}   
\end{figure}

Multimodal training datasets with images and corresponding textual descriptions that include explicit spatial relations tend to be small.  
As a second ingredient of our approach we automatically construct a synthetic dataset with spatial relations named Synthetic Spatial Training Dataset (SSTD), which is used to teach LMs on how to relate location tokens and explicit spatial descriptions. Given an image in an existing dataset, an object detector is used to produce textual descriptions with object labels and location tokens. Given two objects and their bounding boxes, simple rules and templates are used to generate a positive or negative question about the spatial relation between the two objects (or alternatively, about a single object). Figure \ref{fig:std-example} shows such a generated example. The most important advantages of SSTD are: i) it can generate thousands and thousands of different examples, ii) it involves light human labour\footnote{We spent $\sim 5$ hours of work for our specific implementation including rules and templates.}, iii) it can be easily extended to support new spatial relations, and iv) it can be used as a visio-linguistic or text-only dataset. 

To build SSTD, we use the 2014 version of the COCO dataset \citep{lin2014microsoft}. We obtain SSTD training examples from the train set and validation examples from the validation set. Instead of using human annotated object detections, we use automatic VinVL detections, because the vocabulary size of VinVL is much larger than COCO (1848 classes against 80). In COCO, for example, we have the class "person", while VinVL detects more specific classes like "woman", "man", "boy" or "girl", among others, which add more diversity to SSTD. Although VinVL introduces errors in the object detection label or bounding box, this is not important for the text-only case, as we do not need matching visual and textual representations of the image. We are just interested in generating correct spatial relations for the detected object bounding boxes and labels. 

In order to generate SSTD, we manually define a list of interesting and unambiguous spatial relations based on previous work \citep{johnson2018image}. For example, given two bounding boxes, deciding whether an object is \emph{left of} another object, is unambiguous. However, using only those bounding boxes, it is not possible to decide whether the objects are \emph{close to} each other. Even though both BBs may be close, one of the objects can actually be very far in the depth dimension, so we need the context of the image to decide about the spatial relation. In that sense, notice that we did not have to adapt SSTD relations to VSR, just focus on what kind of relations we could unambiguously derive from bounding boxes. In consequence, SSTD should be useful for other tasks involving spatial grounding, not only VSR. In Table \ref{tab:std-relations} we provide all the implemented relations and the category they belong to. All of them can be implemented following some simple rules based on object bounding boxes (more details in Appendix \ref{sec:appendix-sstd}). This is the process we follow to generate an example for SSTD:
\begin{enumerate}
    \item We take an image and check the number of detected objects. As we implement one- or two-object relations, depending on the number of detections, we randomly select among the three categories of Table \ref{tab:std-relations} (i.e. if we have only one detection, we select "object position in the image"). If we have two or more objects, we prioritize two-object relations (i.e, we assign $70\%$ of probability to two-object relations and $30\%$ to one-object relations). Given the category, we randomly sample the required objects (one or two depending on the relation) from all the detections. 
    \item We randomly decide between generating an affirmative or negative question. This way, we make sure that \emph{yes} and \emph{no} answers will be balanced. Using hand-designed verbalization templates, we generate the question corresponding to the spatial relation selected in the previous step (templates are provided in Appendix \ref{sec:appendix-sstd}). 
    \item We verbalize the scene in the image. We provide two kinds of verbalizations: i) generate the textual scene description as the concatenation of all objects detected by VinVL in the image, where each object is accompanied by its location; ii) use only the concatenation of object names, excluding location tokens. Notice that other image verbalization approaches could easily be added, such as captions\footnote{We consider that for our experiments, those alternative verbalization approaches are not interesting, since we want to test how explicit spatial relations are grounded to location tokens.}.
    \item A SSTD example is comprised by a question, a textual scene description and an answer. The image is discarded in the text-only version.
\end{enumerate}

Following this procedure, we can generate many examples from each image. In that sense, SSTD does not have a fixed size: users can decide how many examples they want to extract from each image. In our case, during the spatial training phase of our models, we decide to produce random examples from the same images (COCO train set) in each epoch. That means that the models see an estimate of $num\_epochs \times 80K$ examples during the training process, where $80K$ corresponds to the number of images for COCO train set. Finally, as VSR is also based on COCO, to avoid any contamination, we do not include in the train set of SSTD the images that are already in VSR dev or test splits.

\section{Experiments and results}
We use the \emph{random} split of the VSR dataset for the experiments, given its bigger size. For all the fine-tuning processes described, we train the models in the train set and select the best performing model in the validation set. That model is then evaluated in the test set. Following the recommendations of VSR authors, we provide the average results of three different runs, with the observed standard deviation. The hyperparameters of different models and GPU usage are specified in Appendix \ref{sec:appendix-hyperparameters}.

\subsection{The influence of the location tokens and spatial training}
\label{sec:influence}

We want to assess the importance of two fundamental factors of our approach: i) the use of location tokens for LMs, and ii) the benefits of a spatial training phase using SSTD to better leverage those location tokens. For that purpose, we use BERT-base \citep{devlin2018bert} as our LM and train it in different ways, testing different combinations of using (or not) location tokens and previously training (or not) spatially with SSTD. We add a classification head on top of the \emph{[CLS]} embedding ($\mathbf{t}^{(n_l)}_1$, where $n_l$ is the index of the top layer) for binary classification. We define the head as a multilayer perceptron (MLP) of one hidden layer. We define our MLP in Eq. \ref{eq:mlp}.

\begin{equation}
    \begin{split}
    \mathbf{h} &= \mathrm{LayerNorm}(\mathrm{GELU}(\mathbf{W}_h \mathbf{t}^{(n_l)}_1 + \mathbf{b}_h)) \\
    \mathbf{\hat{y}} &= \mathrm{Sigmoid}(\mathbf{W}_{\hat{y}} \mathbf{h} + \mathbf{b}_{\hat{y}})
    \end{split}
    \label{eq:mlp}
\end{equation}

In order to develop the spatial training phase using SSTD, we randomly built a validation set for SSTD (comprising 40,504 examples) and chose the model which performs best as the one to be used in the VSR experiments.

Table \ref{tab:loc-vs-noloc} shows the obtained VSR test results for the mentioned combinations. The first block shows the performance of BERT-base fine-tuned on the VSR training set, with no significant differences between using or not location tokens. However, we do observe important differences in the second block, where both BERT-base models are previously trained on our Synthetic Spatial Training Dataset (SSTD) and only the model which uses location tokens improves over the previous models. The improvement with the use of spatial training and locations with respect to the other three options is notable, with $\sim 12$ absolute point improvement. 
The results show that location tokens are a good way to encode spatial information for language grounding, and that the spatial training step using SSTD is crucial to make the model learn how that grounding should be done.

On the other hand, Table \ref{tab:spatial-training} shows the results obtained in the validation split of SSTD. Although SSTD is used for spatial training and the obtained results are not the focus of this paper, it is interesting to see how using location tokens, the LM can achieve 94.49 of accuracy, whereas without location tokens, it cannot reach an accuracy of 77. The gap is of around 17 absolute points, which, once again, shows the importance of location tokens.

\begin{table*}[t]
\begin{minipage}{.39\textwidth}
    \centering
    \begin{tabular}{ccc}
    \toprule
     \textbf{Model} & \textbf{Locations} & \textbf{SSTD}\\
    \midrule
    BERT-base & No & 76.96 \\
    BERT-base & Yes & 94.49 \\
    \bottomrule
    \end{tabular}
    \caption{Results (accuracy) on the validation set of our synthetic SSTD dataset. 
    }
    \label{tab:spatial-training}
\end{minipage}
\hfill
\begin{minipage}{.59\textwidth}
    \centering
    \begin{tabular}{cccc}
    \toprule
     & \textbf{Model} & \textbf{Locations} & \textbf{VSR acc}\\
    \midrule
    \multirow{2}{*}{Language Models} & BERT-base & No & 62.11\scriptsize{$\pm$0.88} \\
        & BERT-base & Yes & 61.60\scriptsize{$\pm$0.92} \\
    \midrule
    \multirow{2}{*}{\shortstack{Spatially trained\\ Language Models}} & BERT-base & No & 61.83\scriptsize{$\pm$0.28} \\
        & BERT-base & Yes & \textbf{73.69\scriptsize{$\pm$0.88}} \\
    \bottomrule  
    \end{tabular}
    \caption{Test results on VSR as mean accuracy with standard deviation. First block for language models with and without location tokens. Second block for spatially trained language models (using SSTD) which are then fine-tuned on the VSR training set. 
    }
    \label{tab:loc-vs-noloc}
\end{minipage}%
\end{table*}

\subsection{Comparison with the state of the art}
\label{sec:sota}

In this section we compare our results to the current state-of-the-art models for VSR, and, in addition, we explore whether scaling up LMs brings some extra performance. For that purpose, we use BERT-large as our LM (also adding a binary classification head as in Eq. \ref{eq:mlp}), but we also explore the T5 family of encoder-decoder models \citep{raffel2020exploring}. We include T5 models because the larger size of some models and in order to explore encoder-decoder models, as opposed to encoder-only models such as BERT. To use T5, we add text prefixes before each sentence, such as \textit{'caption:'} for the VSR caption and \textit{'context:'} for the textual scene description. This is done to mimic the input prompts used during the pretraining process of the T5 model, and help the LM to better leverage what it has learnt before. As T5 is a generative LM, it produces answers in an open-ended text generation manner. We select the answer (yes or no) with maximum probability. Thus we do not use any classifier head in this case.

\begin{table*}
\centering
\begin{tabular}{cccc}
\toprule
 & \textbf{Model} & \textbf{Parameters} & \textbf{VSR acc}\\
\midrule
\multirow{4}{*}{\shortstack{Multimodal\\ Systems}} 
    & CLIP \scriptsize{(w/ prompting)} & 632M & 55.2\scriptsize{$\pm$1.4}\\
    & VisualBert\textdagger & 110M & 57.4\scriptsize{$\pm$0.9}\\    
    & ViLT & 87.4M & 69.3\scriptsize{$\pm$0.9} \\
    & LXMERT & 240M & 70.1\scriptsize{$\pm$0.9}\\
\midrule
\multirow{5}{*}{\shortstack{Our\\Spatially trained\\ Language Models}} & BERT-base & 110M & 73.69\scriptsize{$\pm$0.88} \\
    & BERT-large & 336M & \textbf{74.44\scriptsize{$\pm$0.73}} \\
    & T5-base & 220M & 73.09\scriptsize{$\pm$0.59}\\
    & T5-large & 770M & \textbf{74.49\scriptsize{$\pm$0.36}}\\
    & T5-3B & 3B & \textbf{74.52\scriptsize{$\pm$0.25}} \\
\bottomrule   
\end{tabular}
\caption{Test results on VSR as mean accuracy with standard deviation.  
First block for multimodal systems, see text for references. 
\textdagger \ for models with no spatial information. Second block for our spatially trained language models. 
}
\label{tab:sota}
\end{table*}

Table \ref{tab:sota} shows the obtained results for those experiments\footnote{The results of VLMs are directly extracted from \citep{liu2022visual}.}. The best VLM, i.e. LXMERT, obtains an accuracy of 70.1. All our spatially trained LMs surpass that accuracy significantly, which is notable as our models only access bounding box labels and locations, losing potentially important information in the image. The best models are our three largest LMs, with over 74 accuracy, 4 absolute points ahead of LXMERT.

From those results, we can conclude that location tokens and the spatial training phase are good strategies to ground spatial relations in LMs. More importantly, LMs can handle spatial information, which opens the door for applications such as document layout tasks or textual spatial reasoning, for example. However, if we look at the benefits of scaling up the LMs, our experiments show diminishing returns for this specific task. It is true that our best model is a T5 of 3B parameters, however the difference with T5-large or BERT-large is quite small. Notice, though, that we did not perform any extensive hyperparameter tuning, so it could be the case that those larger LMs could actually perform better. Regarding sizes, we would like to note that we use the decoder part of T5 to generate one of Yes or No, and as such it would seem that the decoder is oversized. 

\section{Analysis of the results}

In this section we analyse the results of individual spatial relations, we compare our systems with a rule-based baseline and a VLM, and we finally analyse the use of object attributes.
\subsection{Analysis per spatial relation}
\label{sec:analysis-relation}
\begin{figure*}[t]
\centering
\hspace*{-2cm}
\includegraphics[width=\textwidth]{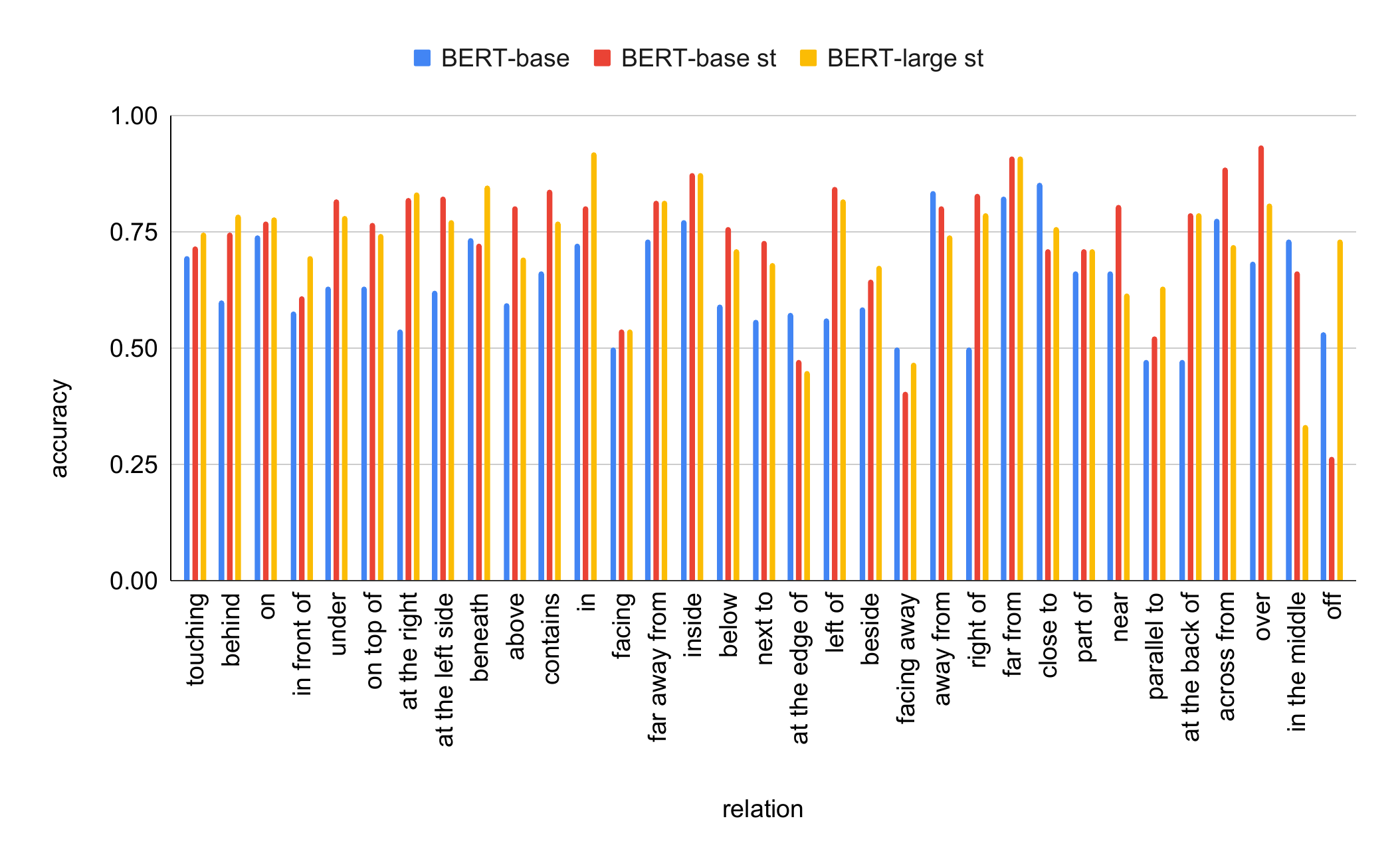}
\caption{Comparison of three BERT models in terms of accuracy per spatial relation. Relations are ordered by frequency in descending order. For readability, we only show the relations that appear more than 15 times in the test set. All three models use location tokens. The "st" acronym in the model name indicates that the model has been spatially trained before the fine-tuning on VSR. Best viewed in color.} \label{fig:model-comp}
\end{figure*}

As the 65 relations in VSR are of different nature, we compare the performance of our spatially trained LMs relation by relation. The objective is to see how spatial training and scale affect the performance. Figure \ref{fig:model-comp} shows the accuracy of three LMs per relation. The selected models are BERT-base with location tokens but without any spatial training, the same BERT-base with spatial training and BERT-large, also with location tokens and spatial training. We only visualize the relations that appear 15 times or more in the test set. 

In general, spatial training helps in almost all relations, with some exceptions. For instance, an orientation relation (\emph{facing away}) and an adjacency relation (\emph{at the edge of}). This could be expected, as SSTD does not cover those relations, because orientation cannot be inferred from BB information, and the object detector in use (VinVL) does not codify it in the attributes either. Orientation seems to be also difficult for VLMs \citep{liu2022visual}, so more work is needed in this regard.

There are also positive effects which show the generalization capabilities of the LMs to some extent. BBs do not provide any 3D information, so we did not include relations like \emph{behind}, \emph{in front of} and \emph{at the back of} in SSTD, but spatial training performs very well for those relations. One of our hypothesis is that SSTD does include size relations (\emph{wider}, \emph{smaller} and so on), and thus the spatially trained models learn to combine BB information with typical size relations to infer depth (e.g. as persons are larger than cats, if a particular person is smaller than a cat, it has to be farther in the scene). We plan to further investigate those cases, since they provide hints of how spatially trained LMs can leverage location tokens to generalize to spatial relations that cannot be described unambiguously with arithmetic rules. We provide a preliminary qualitative analysis in Appendix \ref{sec:appendix-qual-analysis}.

We also observe in Figure \ref{fig:model-comp} that, in general, the performance for VSR relations covered in SSTD (\emph{at the right side of}, \emph{at the left side of}, \emph{on top of}, \emph{above} and so on) improves significantly. Knowledge transfer for those relations was expected, as they are semantically very similar to some SSTD relations. However, in one case, \emph{beneath}, which is tightly related to the SSTD relation \emph{below}, spatially trained BERT-base does not outperform BERT-base, but BERT-large does ($+12$ absolute points). 

To add more context to this analysis, Table \ref{tab:vsr-std-analysis} provides the number of VSR relations per category, alongside the coverage in SSTD and the performance difference between a BERT-base model with and without spatial training (both with location tokens). Overall, SSTD covers only 17 out of the 65 relations in VSR, but there are some relations in SSTD which can be helpful for some other relations in VSR. For example, the VSR relation \emph{detached to} is related to the SSTD relation \emph{overlapping}. Depending on the image, overlapping BBs can be detached objects, but in general, BBs that do not overlap will be detached. Looking at the performance difference (3rd column of Table \ref{tab:vsr-std-analysis}), we can see that spatial training is beneficial for all the categories, except for \textit{topological}, where the difference is very small in any case. The \textit{unallocated} category has an impressive performance gain ($+56.8$), but it is not very significant since there are only 51 examples in the test set. In general, we can say that those categories that are better represented in SSTD, consistently improve in VSR. That is the case of \textit{projective} ($+14.4$), \textit{adjacency} ($+4.7$) and \textit{directional} ($+2.9$). In that sense, the performance gain of $9.1$ absolute points for \textit{orientation} relations is quite surprising.

\begin{table*}
\centering
\begin{tabular}{cccc}
\toprule
 \textbf{VSR category} & \textbf{VSR Relations} & \textbf{In SSTD} & \textbf{Perf. gap}\\
\midrule
Adjacency & 10 & 2 & +4.7\\
Directional & 11 & 2 & +2.9 \\
Orientation & 4 & 0 & +9.1 \\
Projective & 12 & 8 & +14.4 \\
Proximity & 5 & 0 & +1.1 \\
Topological & 18 & 5 & -1.2 \\
Unallocated & 5 & 0 & +56.8 \\
\bottomrule
\end{tabular}
\caption{For every category in VSR, we show how many relations there are. In the second column, we show how many relations are already covered in SSTD. In the last column, the average performance difference between a spatially trained BERT-base against a BERT-base without spatial training is shown.}
\label{tab:vsr-std-analysis}
\end{table*}

Finally, in terms of LM size, the differences between BERT-base and BERT-large are irregular. In general, BERT-large performs better, but there are some cases where BERT-base outperforms it. We do not observe any remarkable behavior.

\subsection{Comparison with a rule-based baseline}
An interesting question that arises from our results is whether our spatially trained LMs learn more than the heuristic spatial rules represented in SSTD. To answer that question, we implemented a rule-based baseline, using the same spatial rules of SSTD to solve the VSR dataset (implementation details can be found at Appendix \ref{sec:appendix-rule-based}). We found that around $38\%$ of test instances could be solved using our spatial rules. However, due to caption-context object matching failures, only $25\%$ of the instances are actually solved using rules. The obtained accuracy for those instances is $60.7$, clearly below the performance of our spatially trained LMs. Indeed, if we solve randomly all the instances that cannot be solved by rules (around $75\%$ of the test set), we obtain an overall accuracy of $52.4$, whereas our best spatially trained LM has an accuracy of $74.5$.

\begin{figure*}[t]
\centering
\includegraphics[width=12cm]{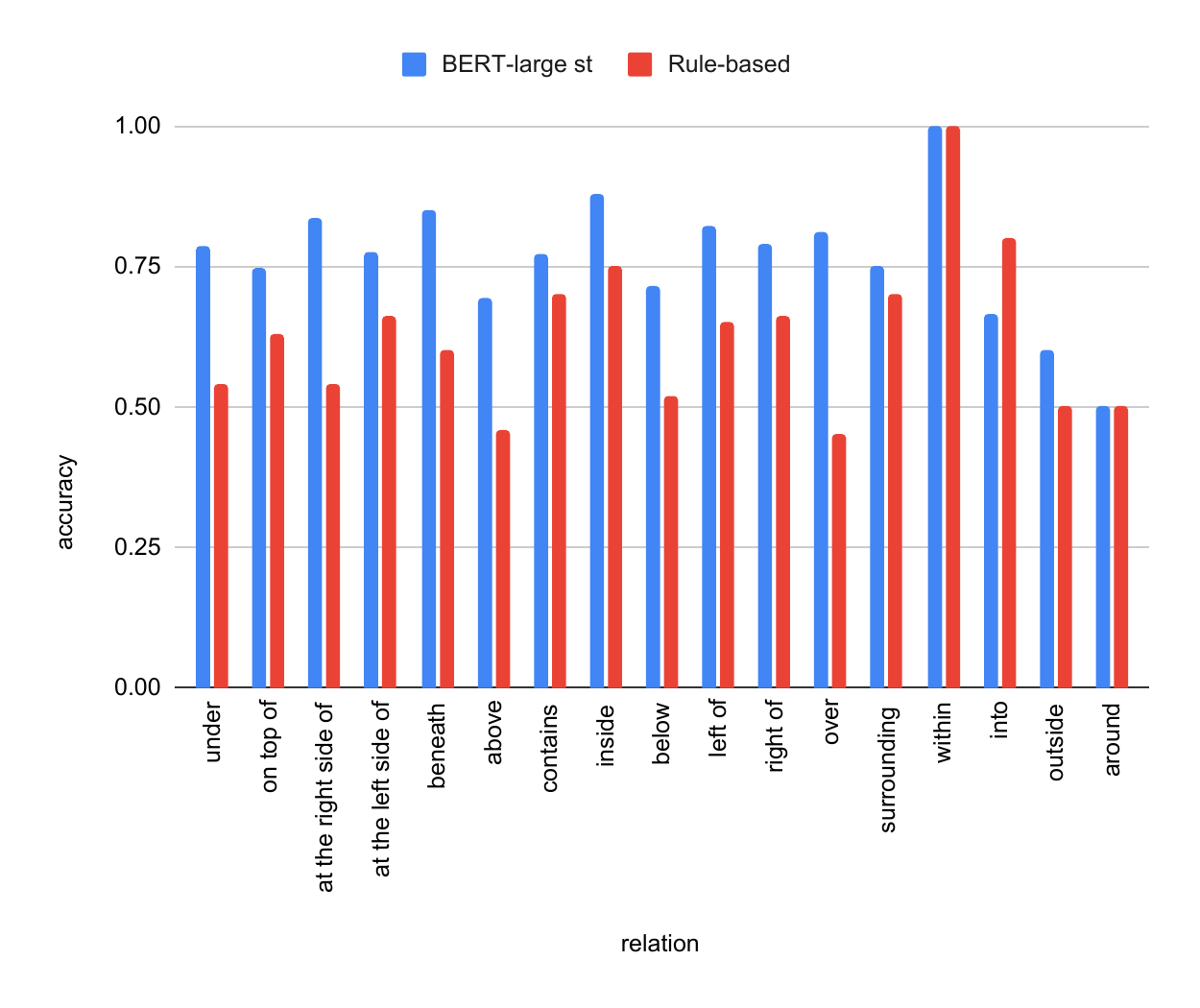}
\caption{Comparison of our spatially trained BERT-large model and the rule-based baseline for the VSR test relations that can be solved using bounding boxes and heuristic rules. Best viewed in color.}\label{fig:rule-vs-learn}
\end{figure*}

Figure \ref{fig:rule-vs-learn} provides a detailed comparison between our rule-based baseline and the spatially trained BERT-large model for VSR test. As can be seen, for all those relations that can be solved using bounding boxes and heuristic rules, the spatially trained LM clearly outperforms the rule-based baseline for all the relations except three: \textit{within} and \textit{around}, where both approaches have the same performance, and \textit{into}, where the rule-based baseline obtains better results (notice, though, that there are only 6 instances for that relation in VSR test, so the results are not very representative). From those results we can conclude that our text-only LMs learn more than the information encoded in the spatial rules of SSTD.

\subsection{Comparison with a VLM}

Even though it is not the main focus of the paper, it is also interesting to see how our spatially trained LMs compare to VLMs. For that analysis, we compare the results of our spatially trained BERT-large and LXMERT for every relation in VSR test.

\begin{figure}
\begin{minipage}{.45\textwidth}
    \centering
    \includegraphics[width=\textwidth]{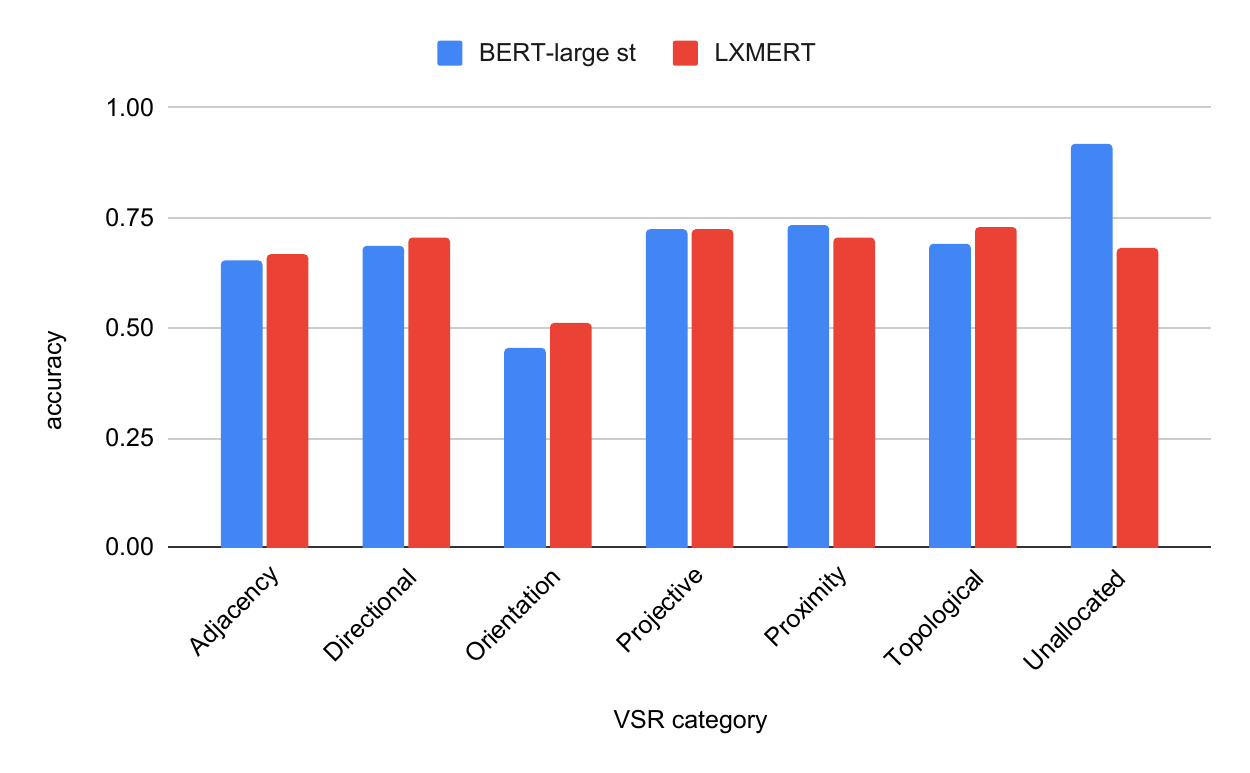}
    \caption{Comparison of our spatially trained BERT-large model and LXMERT for the VSR test categories. Best viewed in color.} \label{fig:lm-vs-vlm}
\end{minipage}%
\hfill
\begin{minipage}{.5\textwidth}
    \centering
    \includegraphics[width=\textwidth]{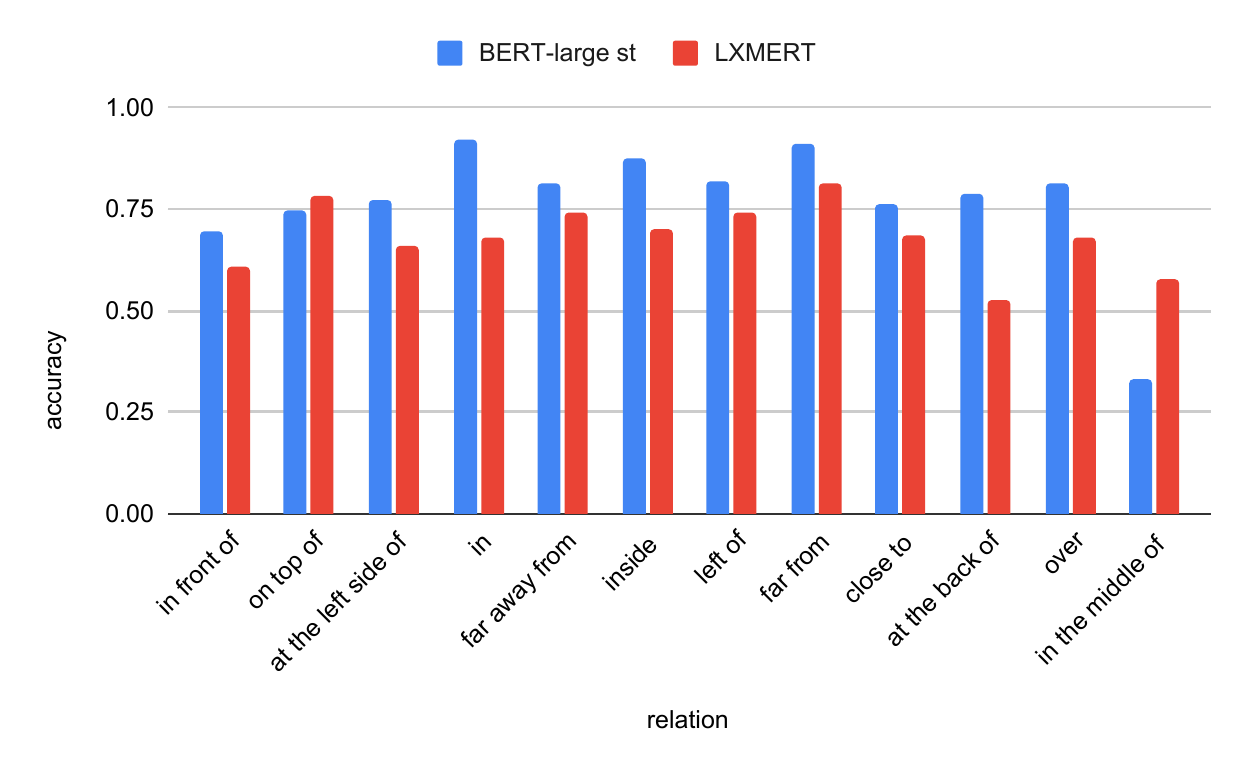}
    \caption{Comparison of our spatially trained BERT-large model and LXMERT for the VSR test relations, where the difference between both models is bigger than 4 absolute points. Best viewed in color.} \label{fig:lm-vs-vlm-rel}
\end{minipage}%
\end{figure}

Figure \ref{fig:lm-vs-vlm} shows the accuracy obtained by both models, grouped by categories. As can be observed, there are no important differences, except for the \textit{unallocated} category, where BERT-large significantly outperforms LXMERT (92 vs 68). However, if we look at the performance relation by relation, there are interesting differences. In Figure \ref{fig:lm-vs-vlm-rel}, we show the accuracy obtained with both models for those relations where the difference is bigger than 4 absolute points (we consider that difference being significant, since it is approximately the overall difference of both models for VSR test). As can be seen, BERT-large outperforms LXMERT for the relations \textit{in front of, at the left side of, in, far away from, inside, left of, far from, close to, at the back of} and \textit{over}. Some relations only require two-dimensional information (\textit{at the left side of, left of, over}) and thus, the better performance of BERT-large could be expected. However, it is curious to see that BERT-large is better than LXMERT for relations like \textit{in front of, in, far away from, inside, far from, close to} and \textit{at the back of}. Those relations should benefit from visual information, but it seems LXMERT cannot leverage that information properly. On the other hand, LXMERT only outperforms BERT-large significantly for the relations \textit{on top of} and \textit{in the middle of}. In the case of \textit{on top of}, the difference is of 4 absolute points and we do not see any clear reason for that difference. For the relation \textit{in the middle of}, BERT-large is specially bad, even worse than BERT-base, which is on par with LXMERT. We believe this behaviour is more related to the low number of instances for that relation in VSR test (only 15).

\subsection{Analysis of the use of object attributes}
VinVL returns not only objects but also their attributes like colors, poses (\emph{open hand}, \emph{standing boy}), sizes, textures (\emph{striped jacket}), materials (\emph{brick wall}) and so on. We modified the spatial training phase to include the attributes in the textual scene description and trained a BERT-base model with the same hyperparameters as in Section \ref{sec:influence}. Afterwards, we fine-tune the best SSTD validation model on the VSR training set. Again, we add the attributes in the textual scene descriptions. The VSR test accuracy is of 74.14, which is inside the standard deviation of the BERT-base models shown in Table \ref{tab:loc-vs-noloc}. We conclude that using object attributes as extracted by VinVL is not beneficial for this specific task, although our analysis in the previous section showed that additional attributes non covered by VinVL like orientation or depth information, if extracted, could be of use.

\section{Conclusions and future work}
\label{sec:conclusions}
In this paper, we have presented a novel way to ground spatial relations in text-only language models through location tokens. To make LMs learn the grounding between spatial relations and location tokens, we also propose the Synthetic Spatial Training Dataset, a textual dataset with unambiguous spatial relations between objects automatically derived from existing images. We run experiments on a verbalized version of the Visual Spatial Reasoning dataset, where spatial grounding can be tested, showing that our approach to ground spatial relations in LMs is effective. Indeed, when compared with VLMs, we obtain even better results, which is another important indication that our spatial grounding approach is working.

Furthermore, scaling up our LMs we obtain the new state-of-the-art in VSR. However, we observe diminishing returns, which may suggest that to ground better those spatial relations, scale is not determinant. That opens the door for other techniques and approaches. 

In the future, we want to deepen on spatial training, including categories like orientation and depth, for example. We also want to transition to text-only spatial reasoning tasks like SpartQA \citep{mirzaee2021spartqa} and RESQ \citep{mirzaee-kordjamshidi-2022-transfer}, where we plan to transform the natural language scene descriptions with explicit spatial relations provided in those tasks, to our textual scene descriptions based on location tokens. We want to see whether those grounded representations do actually improve the spatial reasoning capabilities of LMs.

\section*{Acknowledgments}
Ander is funded by a PhD grant from the Basque Government (PRE\_2021\_2\_0143). This work is partially supported by the Ministry of Science and Innovation of the Spanish Government (AWARE project TED2021-131617B-I00, DeepKnowledge project PID2021-127777OB-C21), and the Basque Government (IXA excellence research group IT1570-22).

\bibliographystyle{unsrtnat}  
\bibliography{spatial-relations-arxiv}

\appendix

\section{SSTD Implementation Details}
\label{sec:appendix-sstd}
We will present here the rules and heuristics followed to derive spatial relations from bounding boxes, grouped by category (see Table 1). We also present the templates we use to generate automatic questions for every case. We assume all BB coordinates are normalized between $[ 0, 1 ]$.
 
 \paragraph{Object position in the image} We first define a region for each of \emph{top left}, \emph{top right}, \emph{bottom left} and \emph{bottom right}. For example, $[0, 0, 0.5, 0.5]$ corresponds to \emph{top left}. If the object BB is inscribed in one of those regions, we return that spatial relation. Otherwise, we check whether the object is in the following regions: \emph{top}, \emph{bottom}, \emph{left} or \emph{right}. An object is in the \emph{left} region, for instance, if the object bounding box is inscribed in the $[0, 0, 0.5, 1]$ region. In all the other cases, the object is in the \emph{center}. Given an object $obj$ and a region $reg$, the template we use for question generation is: "is $\langle obj \rangle$ in $\langle reg \rangle$ region?"
 
 \paragraph{Object size comparison} Assuming two objects $obj_1$ and $obj_2$ and their bounding boxes, we calculate the functions $width(obj)$, $tall(obj)$ and $area(obj)$ for each object, using BB coordinates. If $width(obj_1) > width(obj_2)$, $obj_1$ is \emph{wider} than $obj_2$ (and $obj_2$ is \emph{narrower} than $obj_1$). We apply analogous rules for \emph{taller/shorter} using the $length(obj)$ function and \emph{larger/smaller} using the $area(obj)$ function. Given two objects $obj_1, obj_2$ and a size comparison relation $rel$, the template we use for question generation is: "is $\langle obj_1 \rangle$ $\langle rel \rangle$ than $\langle obj_2 \rangle$?"
 
 \paragraph{Two object positional relations} Assuming two objects $obj_1$ and $obj_2$ and their bounding boxes, if the BB of $obj_1$ is inscribed in the BB of $obj_2$, $obj_1$ is \emph{inside} $obj_2$, and $obj_2$ is \emph{surrounding} $obj_1$. For the relations \emph{left of}, \emph{right of}, \emph{above} and \emph{below}, we use the angle between the centers of both objects. If the center of $obj_2$ is between the angles $[\frac{-3}{4}\pi, \frac{3}{4}\pi]$, we say $obj_2$ is \emph{left of} $obj_1$. Similarly, $[\frac{-3}{4}\pi, \frac{-1}{4}\pi]$ corresponds to \emph{above}, $[\frac{-1}{4}\pi, \frac{1}{4}\pi]$ corresponds to \emph{right of} and $[\frac{1}{4}\pi, \frac{3}{4}\pi]$ corresponds to \emph{below}. Finally, using the Intersection over Union (IoU) of both BBs, we say that $obj_1$ and $obj_2$ are \emph{separated} if their IoU is 0, and \emph{overlapping} if IoU $> 0$. Given two objects $obj_1, obj_2$ and a positional relation $rel$, the template we use for question generation is: "is $\langle obj_1 \rangle$ $\langle rel \rangle$ $\langle obj_2 \rangle$?". In the case of the relation \textit{separated} we use the following template: "are $\langle obj_1 \rangle$ and $\langle obj_2 \rangle$ separated?".

\section{Hyperparameters and GPU Usage}
\label{sec:appendix-hyperparameters}

We always use a grid size $G=32$ all over the experiments. For experiments with BERT-base, both for the spatial training and VSR fine-tuning, we train the models for 20K steps, with AdamW optimizer, a batch size of 56, a maximum learning rate of $5 \times 10^{-5}$, a warmup phase of 2K steps and cosine scheduler for learning rate decay. We use a single NVIDIA A30 GPU to perform all the experiments. Each of the experiments need around 5 hours. 

We train BERT-large models for 20K steps, with a batch size of 32, maximum learning rate of $10^{-5}$, AdamW optimizer, warmup phase of 2K steps and cosine scheduler. Using a NVIDIA A100 GPU, we need around 4 hours for the spatial training and additional 5 hours for fine-tuning on VSR. In the case of T5 we train the models spatially for 88K steps (T5-3B is trained for 20K steps due to its size) and fine-tune on VSR for 20K. We use a batch size of 32, AdamW optimizer, maximum learning rate of $5 \times 10^{-5}$, a warmup phase of 2K steps and cosine scheduler for learning rate decay. Regarding the T5 family: T5-base is trained on 1 NVIDIA A30 GPU: for spatial training it needs $\sim 20$ hours and for VSR fine-tuning $\sim 3.5$ hours. T5-large is trained on 1 NVIDIA A100 GPU: it needs 1 day and $\sim 4$ hours for spatial training, whereas VSR fine-tuning takes $\sim 3.5$ hours. Finally, T5-3B is also trained on a single NVIDIA A100 GPU: spatial training $\sim 20$ hours (20K steps) and VSR fine-tuning $\sim 15$ hours. 

No hyperparameter search was performed.

\section{Qualitative analysis of generalization capabilities}
\label{sec:appendix-qual-analysis}
\begin{figure*}[t]
\centering
\includegraphics[width=15cm]{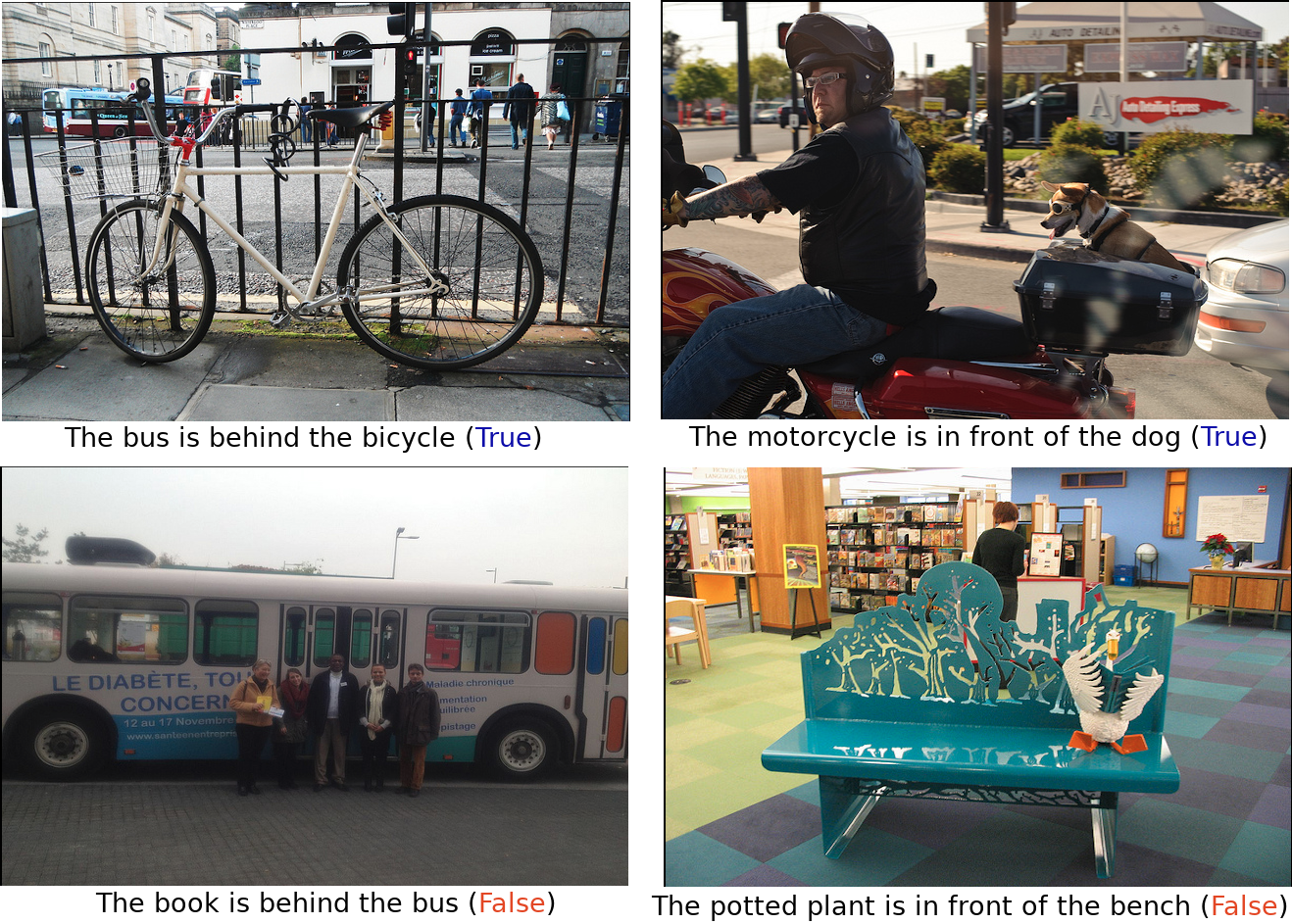}
\caption{Comparison of the predictions of two BERT models for VSR test examples. The spatially trained BERT model predicts correctly the labels, whereas the BERT which has been trained only on VSR does not.} \label{fig:qual-analysis}
\end{figure*}

We compare some examples of two text-only LMs: the BERT-base model with location tokens trained only on VSR (BERT for short) and the BERT-base model with location tokens trained on SSTD and fine-tuned on VSR (st-BERT for short). We want to see the effects of the spatial training on SSTD to better generalize in VSR. For that purpose, we focus on two relations that cannot be represented in SSTD, since they cannot be unambiguously defined with BB information and involve 3D arrangement of objects: \textit{behind} and \textit{in front of}. For \textit{behind}, the accuracy of BERT is 0.6 and the accuracy of st-BERT is 0.75, calculated over 136 examples. For \textit{in front of}, BERT scores 0.58 and st-BERT 0.61 (116 examples). Those results show that SSTD training helps even when the spatial relations are not represented in the dataset. Figure \ref{fig:qual-analysis} offers some intuition of why this might be happening. For the first example, we see that the bus is much smaller than the bike. As SSTD includes relative size relations, we think the model has learned that buses are typically bigger than bikes. Thus while training on VSR, the model might be able to leverage that information and relate size differences with 3D arrangements of objects. A similar reasoning can be applied to the second (motorcycle and dog) and the last examples (bench and potted plant), but for the \textit{in front of} relation. For the third example (bus and book), it seems st-BERT could leverage the fact that the book can only be visible if it is in front of the bus, given the arrangement of the bus BB. However, BERT could not predict the spatial relation correctly. 

We also analyse two other relations that are not in SSTD, but are also related to relative object sizes: \textit{next to} and \textit{far from}. For \textit{next to}, BERT obtains 0.56 and st-BERT 0.73 (over 41 examples). For \textit{far from} BERT scores 0.83 and st-BERT 0.91 (over 23 examples). Notice that the relation \textit{far away from} is very similar to \textit{far from} and st-BERT clearly outperforms BERT also (0.88 vs 0.73 over 49 examples). For the first example (pizza and chair), given the small size of the chair, it can be inferred that it is far in the depth dimension. It seems st-BERT can leverage this information, whereas BERT cannot. For the second example (refrigerator and cat), both BBs overlap and it seems st-BERT infers that situation cannot lead to two objects far away given the typical sizes of those objects. The third example (backpack and cat) shows a case where both BBs are slightly overlapping. Again, the typical sizes of both objects could lead st-BERT to infer that they are actually next to each other. Finally, for the fourth example we see that the hot dog BB is inside the bowl BB. st-BERT infers that this is not the typical arrangement for \textit{next to}, but BERT cannot do that, even though it has the same textual scene representation. 

\begin{figure*}[b]
\centering
\includegraphics[width=15cm]{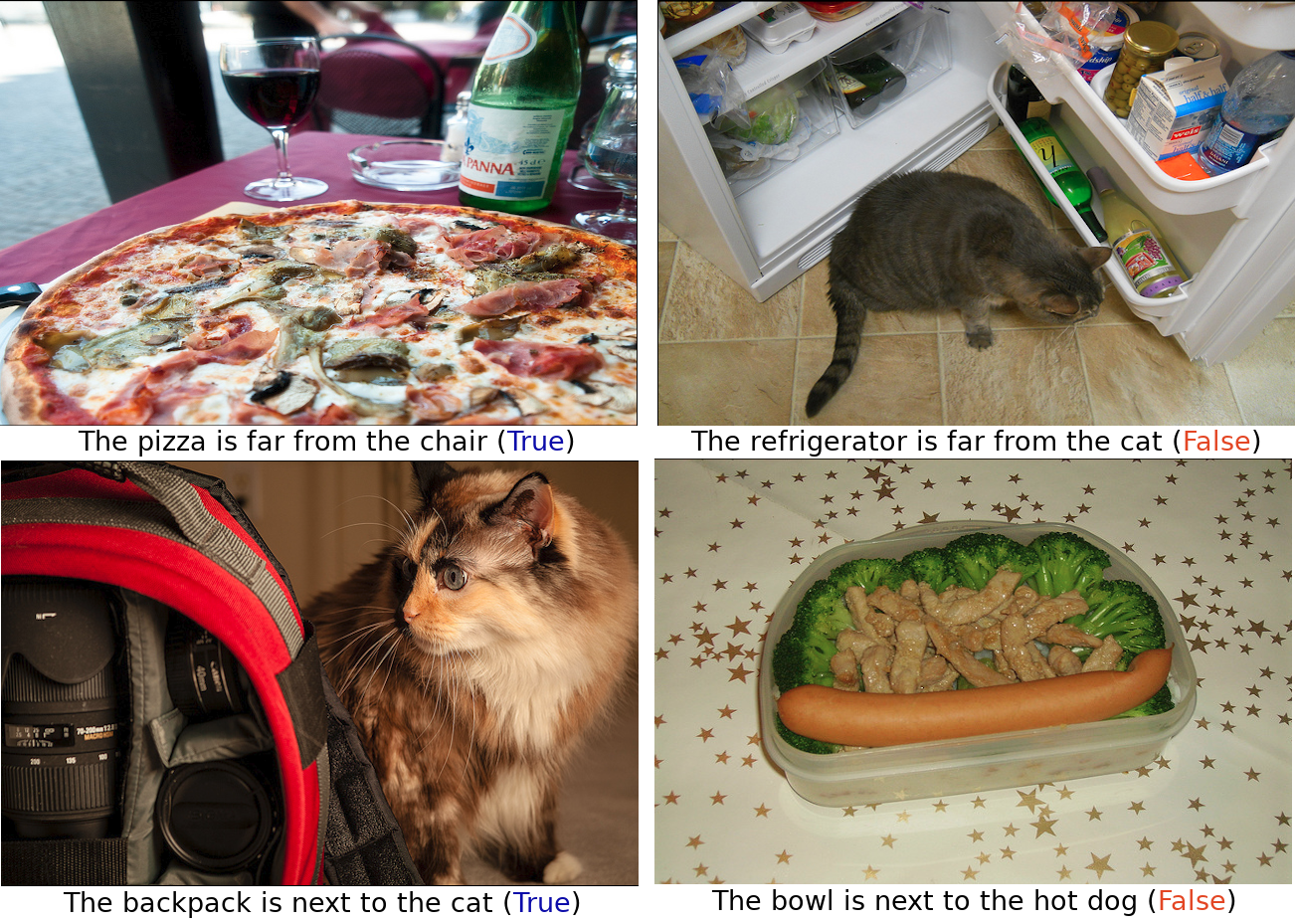}
\caption{Comparison of the predictions of two BERT models for VSR test examples. The spatially trained BERT model predicts correctly the labels, whereas the BERT which has been trained only on VSR does not.} \label{fig:qual-analysis2}
\end{figure*}

\section{Implementation details of the rule-based baseline} 
\label{sec:appendix-rule-based}
To implement the rule-based baseline, we first defined manually a mapping between VSR relations and SSTD relations. As shown in Section \ref{sec:analysis-relation}, only 17 VSR relations out of 65 can be mapped to SSTD relations. That mapping is shown in Table \ref{tab:vsr-sstd-mapping}. Given a VSR test instance, we check the spatial relation (provided in the annotations of the dataset) and if it can be mapped to a SSTD relation, we perform the following steps: a) from the VSR caption, we retrieve the subject and object using string manipulation; b) we find the same subject and object in the textual scene description, using string matching; c) if both subject and object are found, we retrieve their bounding boxes and apply SSTD rules to solve the instance; d) if any of subject or object are not found, or the relation cannot be mapped to a SSTD relation, we choose the answer randomly ($50\%$ of probability).

\begin{table*}
\centering
\begin{tabular}{cc}
\toprule
 \textbf{VSR relation} & \textbf{SSTD Relations}\\
\midrule
at the right side of & right of \\
at the left side of & left of \\
around & surrounding \\
into & inside \\
on top of & above \\
beneath & below \\
left of & left of \\
right of & right of \\
under & below \\
below & below \\
above & above \\
over & above \\
contains & surrounding \\
within & inside \\
surrounding & surrounding \\
inside & inside \\
outside & separated \\
\bottomrule
\end{tabular}
\caption{The mapping between VSR relations and SSTD relations.}
\label{tab:vsr-sstd-mapping}
\end{table*}

\end{document}